%% file: ms.tex
\documentclass[a4paper,11pt]{llncs}

\usepackage{graphics}
\usepackage{graphicx}
\usepackage[margin=1in]{geometry}
\usepackage{amsmath}

\usepackage[round]{natbib}

\begin{document}

\title{An empirical study on the names of points of interest and \\their changes with geographic distance}
\author{Yingjie Hu$^1$\thanks{Corresponding author; email: yjhu.geo@gmail.com \protect\\ To cite this paper: Hu, Y. and Janowicz, K., 2018. An empirical study on the names of points of interest and their changes with geographic distance. In \textit{Proceedings of the 10th International Conference on Geographic Information Science (GIScience 2018)}. Aug. 29-31, Melbourne, Australia.} and Krzysztof Janowicz$^2$}
\institute{
$^1$GSDA Lab, Department of Geography, University of Tennessee, Knoxville, USA \\
$^2$STKO Lab, Department of Geography, University of California, Santa Barbara, USA
}

\maketitle

\input{abstract}

\input{intro.tex}

\input{related.tex}

\input{dataset.tex}

\input{exp.tex}
\input{conclusion.tex}
\bibliographystyle{splncsnat}
\bibliography{references}

\end{document}

%% file: abstract.tex
\begin{abstract}
While Points Of Interest (POIs), such as restaurants, hotels, and barber shops, are part of urban areas irrespective of their specific locations,  the names of these POIs often reveal valuable information related to local culture, landmarks, influential families, figures,  events,  and so on. Place names have long been studied by geographers, e.g., to understand their origins and relations to family names. However, there is a lack of large-scale empirical studies that examine the \textit{localness} of place names and their changes with geographic distance. In addition to enhancing our understanding of the coherence of geographic regions, such empirical studies are also significant for geographic information retrieval where they can inform computational models and improve the accuracy of place name disambiguation. In this work, we conduct an empirical study based on 112,071 POIs in seven US metropolitan areas extracted from an open Yelp dataset. We propose to adopt term frequency and inverse document frequency in geographic contexts to identify local terms used in POI names and to analyze their usages across different POI types. Our results show an uneven usage of local terms across POI types, which is highly consistent among different geographic regions. We also examine the decaying effect of POI name similarity with the increase of distance among POIs. While our analysis focuses on urban POI names, the presented methods can be generalized to other place types as well, such as mountain peaks and streets.  \\


\end{abstract}

%% file: intro.tex
\section{Introduction}

People name the environment that surrounds them to communicate about it.  Almost every aspect of geographic space that can be described and depicted can be named. It has been suggested that place names, or toponyms, play a key role in stabilizing the otherwise unwieldy space into more manageable textual inscriptions \citep{palonen1993reading,kearns2002proclaiming,rose2008sixth}. From a perspective of \textit{space} and \textit{place} \citep{tuan1977space}, the creation of a place name signifies the important moment when  people explicitly integrate human experience with space. 

Place names, made available via digital gazetteers, power GIS,  geographic information retrieval (GIR), and modern search engines and recommender systems more broadly \citep{jones2008geographical,goodchild2008introduction,vasardani2013locating}. After all, people communicate using place names not coordinates. Interestingly, and in difference to human geography, most GIR research simply uses place names as identifiers instead of examining how those names were formed and how similar they are to nearby names. This is understandable since we are often interested in questions such as \textit{What are the best Italian restaurants within 10 minutes driving?} instead of the specific names of these restaurants or what they reveal about the history of a region, such as immigration trends.

Place names have long been studied in human geography with a traditional focus on etymology and place taxonomies \citep{zelinsky1997along,rose2010geographies}. For example, the place name \textit{Las Vegas} means \textit{The Meadows} in Spanish and points to the former abundance of wild grasses and desert springs, both of which were crucial information for travelers and led to the descriptive place name. 
While such studies provide in-depth explanation of place names, they are often limited to case-by-case examinations with qualitative descriptions. This could include studies focusing on specific regions, names, places types, and so forth.

In contrast, this work is based on more than $110,000$ place names of different types distributed across seven metropolitan areas within the US. Our focus is on uncovering term usage patterns and their relations with geographic locations, e.g., as modeled by a decaying influence or local names with increasing distance. There are several reasons for performing such a large-scale, data-driven study. First, place names reveal many social and cultural characteristics, and can help us understand various aspects of urban areas. Previous research in human geography has considered place names, such as street names, as \textit{city-text} embedded in the cityscape \citep{azaryahu1990renaming,azaryahu1996power}. A systematic examination on these city-texts, can help expand our knowledge of the studied regions. Second, large-scale empirical research examining place names can aid in  discovering common principles in place naming and relations to environments. This can be distinguished from case-by-case place name studies in which the discovered knowledge often cannot be generalized to other names or geographic areas. Third, such studies can facilitate the development of computational models for places. We can integrate the discovered common principles, socio-cultural characteristics, and local terms  into computational models, e.g., via an implemented knowledge base, to better support tasks such as place name disambiguation \citep{amitay2004web,leidner2008toponym,overell2008using,hu2014improving}. This last point is a key strength of this work. Our results can act as a quantitative foundation for the localness of identifiers \textit{per place}, which will enable future research to push the envelop on place name disambiguation. In fact, our previous \textit{Things and Strings} place disambiguation method \citep{ju2016things} has demonstrated the usefulness and need for combining geographic and linguistic information.


The names of Points Of Interest (POIs), such as restaurants, hotels, grocery stores, and auto repairs, are examined in this work. These POI names are from an open dataset released by Yelp, a company that provides search services for local businesses. POIs play important roles in supporting many aspects of our daily life \citep{mckenzie2015poi,Novack2017,yan2017itdl}. One reason we select POI names for this study is that these names reflect more of the diverse views of the general public, since the business owners can decide on names themselves. This can be differentiated from other place names, such as street names, which often result from political and administrative decisions \citep{azaryahu1996power,alderman2000street,rose2008number}. In addition, the names of POIs often contain local information, such as city or state names, natural or man-made geographic features, vernacular names, local families (e.g., a family-owned business), language patterns, local cultural differences, and others. Figure 1 shows an example of searching for the word ``Vol'' in the city of Knoxville, Tennessee, USA using Google Maps. It returns many places which use this term as part of their names, as ``Vol'' is the local nickname of the popular football team ``Volunteer''. The use of American sports team names in toponyms was also noted in previous human geography research \citep{baggio2006dawn}. In GIR and place name disambiguation, understanding the link between ``Vol'' and the city of Knoxville can help locate related place names more accurately.
\begin{figure}[h]
	\centering
	\includegraphics[width=0.7\textwidth]{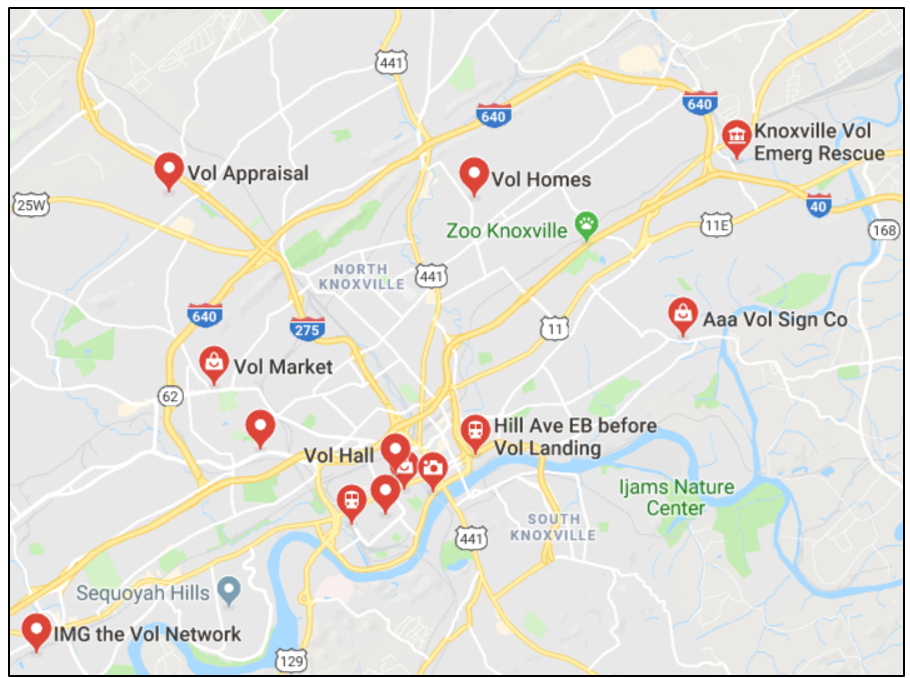}
	\caption{An example of POIs in Knoxville, TN, USA that use ``Vol'' as part of their names.}
\end{figure}


More specifically, we aim to answer the following questions in this work: 1) what are the local terms that are used in POIs in different geographic areas? 2) how are these local terms used in different types of POIs, such as restaurants, hotels, and barber shops? and 3) how do POI names change with geographic distance? \textbf{The contributions of this paper are as follows:}
\begin{itemize}
	\item We propose adopting the technique of term frequency and inverse document frequency in geographic contexts to identify local terms used in POIs in different metropolitan areas.
	\item We find an uneven usage of local terms in the names of POIs across POI types, and such an uneven usage is highly consistent across the seven studied metropolitan areas.
	\item We test two types of models, count-based vector and word2vec, for understanding and capturing the distance decay effect of the similarity of POI names.  
\end{itemize}

The remainder of this paper is structured as follows. Section 2 reviews related work on place names and toponym disambiguation. Section 3 describes the dataset used in this study and an exploratory data analysis. Section 4 presents methods and experiments for identifying local terms from POI names, examining their usages across POI types, and modeling the distance decay effect of POI name similarity. Section 5 summarizes this work and discusses future directions.



%

%% file: related.tex
\section{Related Work}

Place names have attracted the interest of many researchers in geography. For decades, geographers have been collecting and categorizing place names, studying their origins, and understanding their meanings \citep{wright1929study,zelinsky1997along,nash1999irish}. It has  been argued that the act of assigning a name to \textit{space} plays a key role in producing the social construct of \textit{place} \citep{rose2010geographies}. As suggested by \cite{carter1987road}, place names transform space into knowledge that can be read. The social, cultural, and political implications of place names have been widely studied \citep{azaryahu1986street,azaryahu1990renaming}. Examples include the renaming of streets after the establishment of a new regime to memorize new stories \citep{light2004street,rose2008number}, the use of street names to challenge racism  \citep{alderman2002street,alderman2016place}, and assigning more marketable names to local businesses and hospitals \citep{raento2001naming,kearns1999boldly}.


Digital gazetteers provide systematic organizations of place names (N), place types (T), and spatial footprints (F) \citep{hill2000core,goodchild2008introduction}. As valuable knowledge bases, gazetteers provide important functions for various applications by connecting the three core components.  The key functions of a gazetteer include lookup (N $\rightarrow$ F), type-lookup (N $\rightarrow$ T), and reverse-lookup (F($\times$ T) $\rightarrow$ N) \citep{janowicz2008role}. The first case, for example, corresponds to a query for the spatial footprint of the place name \textit{CMS Auto Care}, the second to the place type, and the third to the place names given the spatial footprint and a place type (e.g., \textit{Automotive}). Research was conducted to enrich gazetteers with (vague) place names and their fuzzy spatial footprints. \cite{jones2008modelling}, for instance, used a search engine to harvest geographic entities (e.g., hotels) related to vague place names (e.g., ``Mid-Wales''), and utilized the locations of these harvested entities to construct vague boundaries. Flickr photos present a natural link between textual tags and locations, and have been used in many studies on identifying the boundaries of vague places and regions \citep{grothe2009automated,kessler2009bottom,intagorn2011learning,li2012constructing}. \cite{twaroch2010web} developed a Web-based platform, called ``People's Place Names'', which invites local people to contribute vernacular place names.

In geographic information retrieval \citep{jones2008geographical}, place names are frequently discussed in the context of place name disambiguation. Since different place names can refer to the same place instance and the same place name can refer to different place instances, it is challenging to determine which place instance was referred to by a name in text, e.g., the abstract of a news article  \citep{amitay2004web,leidner2008toponym}. Gazetteers have been used in many ways for supporting place name disambiguation. Based on the related places in a gazetteer (e.g., higher-level administrative units), researchers  developed methods, such as co-occurrence models \citep{overell2008using} and conceptual density \citep{buscaldi2008conceptual}, to disambiguate place names. Based on the spatial footprints of place instances, researchers  designed heuristics  for place name disambiguation, e.g., place names mentioned in the same document generally share the same geographic context  \citep{lieberman2010geotagging,santos2015using}. The process of recognizing and resolving place names from texts is called \textit{geoparsing} \citep{gelernter2011geo,karimzadeh2013geotxt,gritta2017s,wallgrun2018geocorpora}. Place names are also examined in studies on toponym matching and geo-data conflation \citep{santos2018toponym}.

Few existing studies, however, have empirically examined the term usage of place names and their relations with geographic locations based on large datasets. \cite{longley2011creating} and \cite{cheshire2012identifying} investigated the geospatial distributions of surnames based on the data from the Electoral Register for Great Britain and delineated surname regions. Their study is related to our work, since family names are included in the names of some local business. We perform an empirical study based on a large number of POI names in different US metropolitan areas. Compared with the existing literature, this work is unique in that it examines the local terms in POI names, explores the term usage patterns, and analyzes the relations of POI names to geographic locations as well as their decay in this relationship over distance. 

%% file: dataset.tex
\section{Dataset}
\begin{figure}[h]
	\centering
	\includegraphics[width=0.95\textwidth]{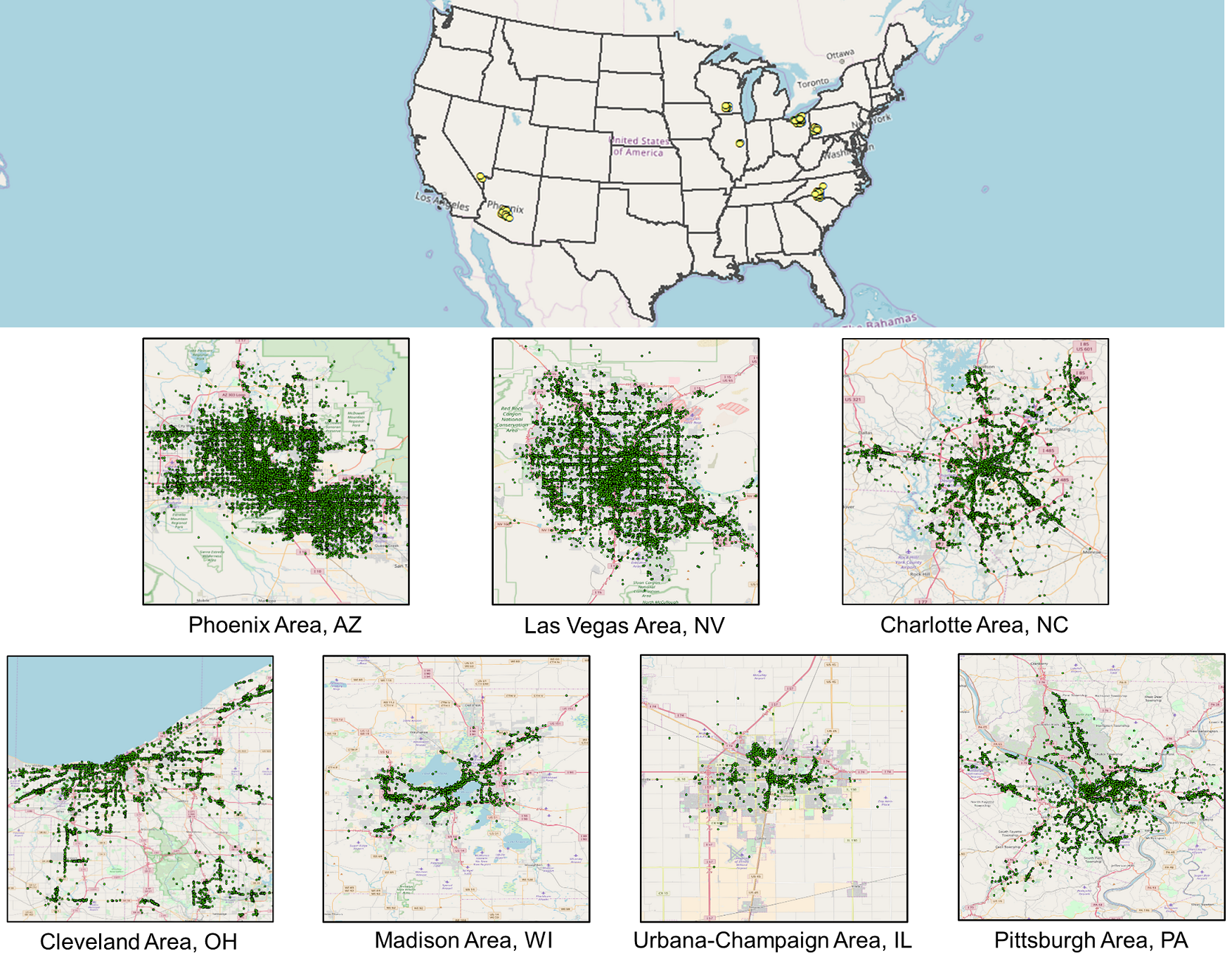}
	\caption{The seven US metropolitan areas and their POIs used for this study.} \label{POI_city}
\end{figure}

We first describe the data used in this empirical study, which is an open POI dataset from Yelp (\url{https://www.yelp.com/dataset}). The original dataset contains POIs from 11 metropolitan areas in four countries: the US, Canada, the UK, and Germany. Considering the language differences in POI names (e.g., German and English) and the barrier effects of country borders, we focus on the seven metropolitan areas within the US, which contain $112,071$ POIs. Each POI data record has the POI name, city name, state name, latitude-longitude coordinates, and other information, such as the number of reviews and average rating. Figure \ref{POI_city} shows the general locations of the seven metropolitan areas and the geographic distributions of the POIs in each of these areas.


We start by performing an exploratory analysis on the term usage frequency in the POI names. It has been found that Zipf's law exists in the usage of terms in natural language texts \citep{manning1999foundations},  namely the frequency of a term is proportional to the inverse of its frequency rank among all terms (Equation \ref{zipf}).
\begin{equation}
 f  \propto \frac{1}{r}   \label{zipf}
\end{equation}
where $f$ is the frequency of a term and $r$ is the rank of the term among all terms based on frequency. According to Zipf's law, a small number of terms are used highly frequently while most others are  used only occasionally.  The names of POIs are different from natural language texts in that they are typically not complete sentences but phrases. In this situation, does Zipf's law still hold in POI names?

\begin{figure}[h]
	\centering
	\includegraphics[width=\textwidth]{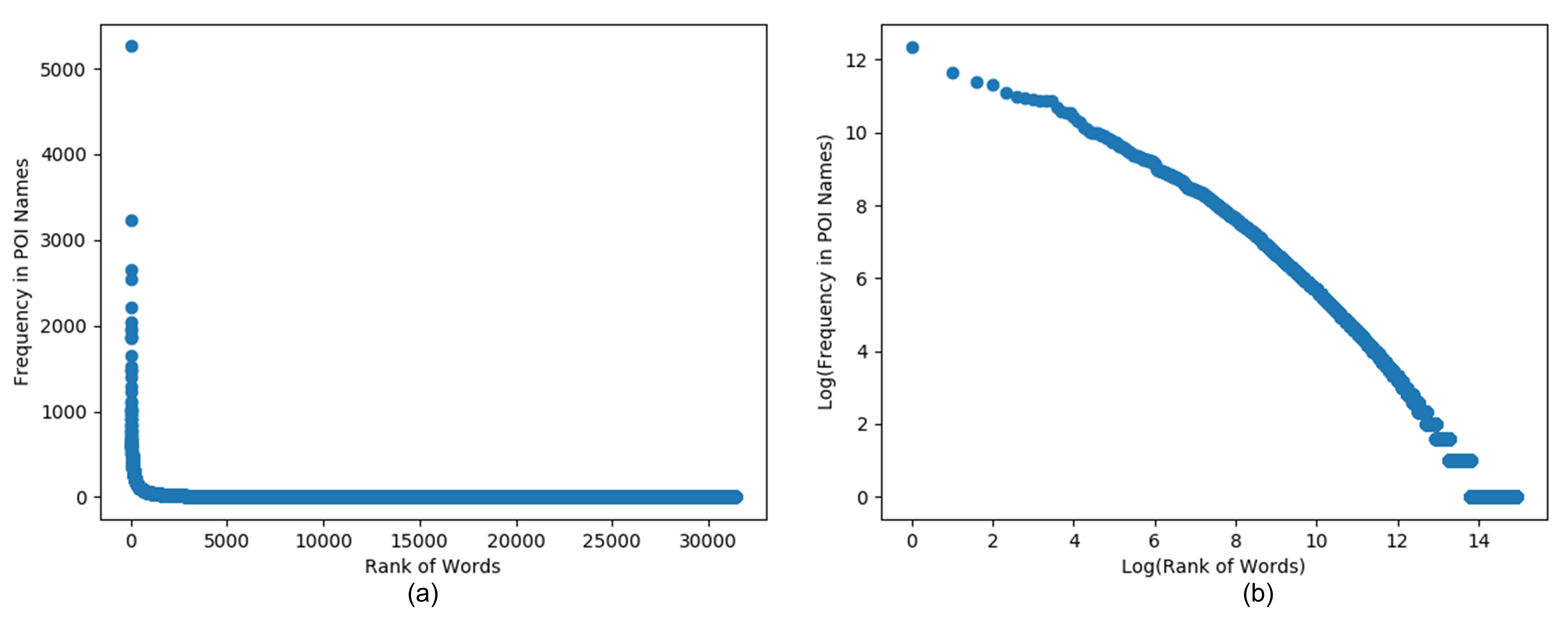}
	\caption{Term frequencies and their ranks in POI names: (a) original values; (b) log-log plot.} \label{word_usage}
\end{figure}

To answer this question, we develop a Python script\footnote{Source code is available at: \url{https://github.com/YingjieHu/POI\_Name}} which reads through the names of the POIs in the seven metropolitan areas, counts the frequencies of all terms contained in each name, and ranks the terms based on their frequencies. We then use the ranks as the horizontal coordinates and term frequencies as the vertical coordinates, and the result is shown in Figure \ref{word_usage}(a).
As can be seen, there is a highly skewed distribution of term frequency with a long tail, which suggests that a small number of terms are used much more frequently  than most other terms. In fact, Figure \ref{word_usage}(a) shows almost a right angle fall-off  since the term frequency decreases rapidly with a small increase of the rank. The log-log plot of the frequencies and ranks is shown in Figure \ref{word_usage}(b), and we see almost a straight line. To quantitatively measure the match of term usage in POI names to Zipf's law, we fit a linear regression model with $\log f = A + b \log r$, and obtained an R-squared value of $0.962$. Based on this exploratory analysis, we conclude that the term usage in POI names also follow Zipf's law, even though POI names are usually not complete sentences. The top $10$ most frequent terms in POI names in this Yelp dataset are: \textit{the}, \textit{and}, \textit{of}, \textit{center}, \textit{pizza}, \textit{grill}, \textit{spa}, \textit{bar}, \textit{auto}, \textit{restaurant}. These most frequent terms reflect the inherent characteristics of POI names and POI types. It is worth noting that the most frequent terms in POI names may change across countries, depending on the corresponding cultures and lifestyles.




%% file: exp.tex
\section{Data Analysis}
In this section, we perform in-depth analyses on POI names. We organize this section into three subsections based on the three core components of gazetteers \citep{hill2000core}. Thus, the first subsection focuses on \textit{place names}, and aims to identify the local-specific terms used in these POI names. The second subsection looks into the interaction between POI names and \textit{place types}, and examines the usage of local terms in different POI types. Finally, the third subsection analyzes the change of POI names with geographic distance based on the \textit{spatial footprints} of the POIs. 

\subsection{Identifying local terms from POI names}
In this first analysis, we attempt to answer the question: \textit{what are the local terms used in the names of POIs in a geographic area?} While not every POI name contains local specific terms, some names are influenced by local factors, such as the ``Vol'' example discussed in the Introduction. We consider local terms as those frequently used in a local geographic area but less likely to be used in other areas. Identifying these local terms can help enhance computational models for place name disambiguation. We make use of the technique, term frequency and inverse document frequency (TF-IDF), a method commonly used in information retrieval, and adapt it to the context of geography. Equation \ref{tf-idf} shows the adapted version of TF-IDF.
\begin{equation}
w_{ij} = \mathit{tf}_{ij} \times \log\frac{|G|}{|G_j|}  \label{tf-idf}
\end{equation} 
where $w_{ij}$ is the weight of a term $j$ in geographic area $i$, $\mathit{tf}_{ij}$ is the frequency of term $j$ in area $i$, $|G|$ is the total number of geographic areas in a study (which is seven in our case), and $|G_j|$ is the number of geographic areas that contain the term $j$. TF-IDF will highlight the terms that are frequently used in a local area, while reducing the weights of those commonly exist in POI names everywhere. In fact, the weights of the terms that occur in all seven metropolitan areas will become zero based on Equation \ref{tf-idf}.

\begin{figure}[h]
	\centering
	\includegraphics[width=\textwidth]{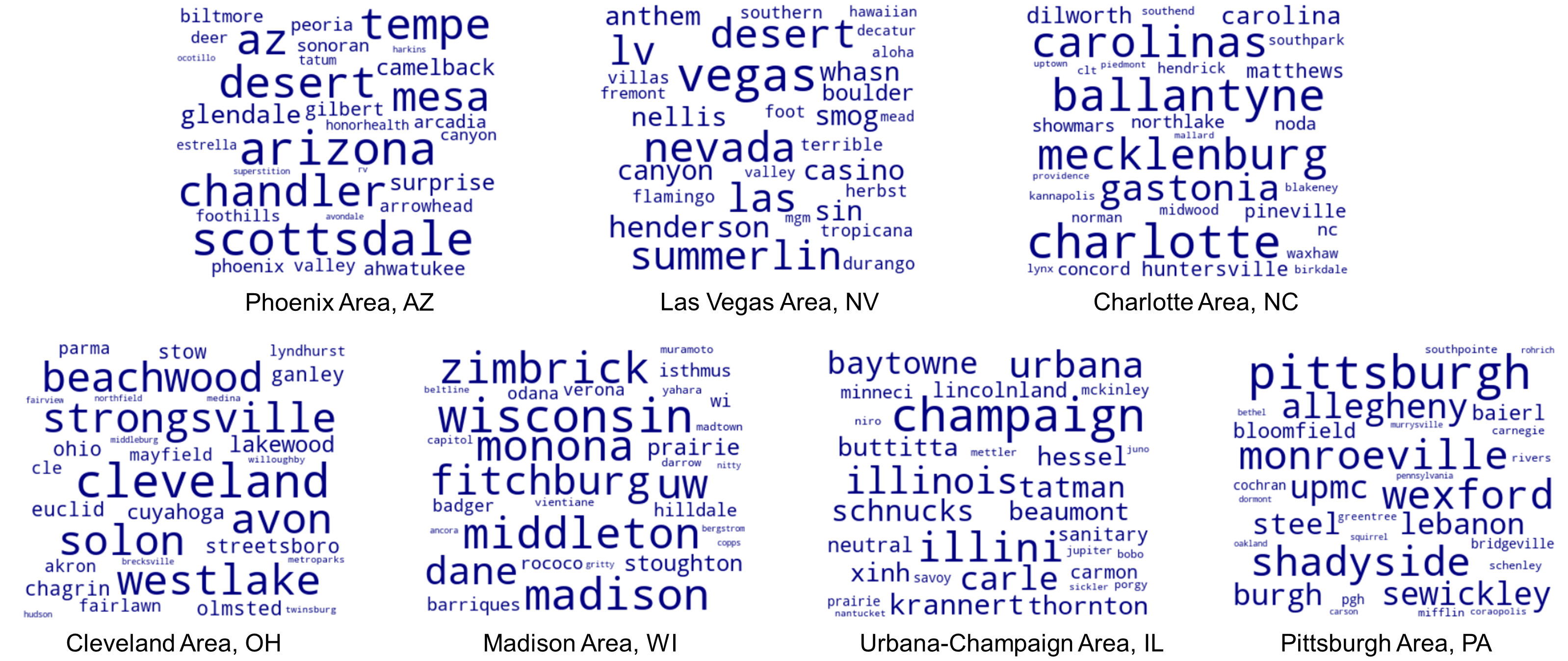}
	\caption{Local terms identified based on the POI names in the seven US metropolitan areas.} \label{word_cloud}
\end{figure}

Before applying the adapted TF-IDF to the POI names, we perform several data pre-processing steps. All POI names are converted to lowercase, and punctuations in POI names are removed. We did not remove typical stop words,  such as ``the'' and ``of'', since the term frequencies in POI names are not the same as other natural language texts, as shown in the exploratory analysis. Thus, typical stop words may not be stop words in the names of POIs. We also performed one special step for this analysis by counting the exact same POI names only once within a metropolitan area. The rationale behind this step is that term frequency can be increased in two situations: 1) one term is used by many different POIs (e.g., the term ``Vol'' is used in the names of many POIs); and 2) one word is used by the same POI business which simply shows up many times in a metropolitan area (e.g., ``walmart''). We would prefer to keep the terms in the first situation, since those are endorsed by many different POIs and are more likely to be valid local terms than those in the second situation. After removing these repeating POI names, we group the names that belong to the same metropolitan areas using the bag-of-words model. We then use the adapted TF-IDF to identify local terms. Figure \ref{word_cloud} shows the top $30$ local terms identified for each of the seven metropolitan areas.

We can group the identified local terms into the following categories:
\begin{itemize}
	\item \textbf{City names}: This is the most common type. POI names in all seven metropolitan areas contain city names, such as \textit{scottsdale}, \textit{las vegas}, \textit{charlotte}, and \textit{cleveland}.
	\item \textbf{State names}: This is similar to city names. State names, such as \textit{arizona} and \textit{wisconsin}, are used in POI names. There are also name abbreviations, such as \textit{az} and \textit{wi}.
	\item \textbf{Natural features}: Examples include \textit{desert} and \textit{canyon} in Phoenix and Las Vegas areas, \textit{prairie} in Madision and Urbana-Champaign areas, and \textit{rivers} in Pittsburgh area.  
	\item \textbf{Sports teams}: Examples include \textit{badger} in Wisconsin and \textit{illini} in Illinois. 
	\item \textbf{Family names}: A notable example is \textit{zimbrick} in Madison, Wisconsin,  a regional car dealer started by \textit{John Zimbrick} (\url{http://www.zimbrickbuickgmceast.com/Zimbrick-History}). 
	\item \textbf{Local cultures}: Terms such as \textit{sin} and \textit{casino} are observed in the POI names in Las Vegas, while the term \textit{steel} is observed in the POI names in Pittsburgh area. 
\end{itemize}

\subsection{Examining local term usage in different POI types}
The first analysis identified the local terms used in POI names in each geographic area. However, do POIs in different types have similar probabilities in using local terms as part of their names? In addition, are there  regional differences in using local terms for names among POI types? In this second analysis, we aim to answer these questions.

\begin{figure}[h]
	\centering
	\includegraphics[width=\textwidth]{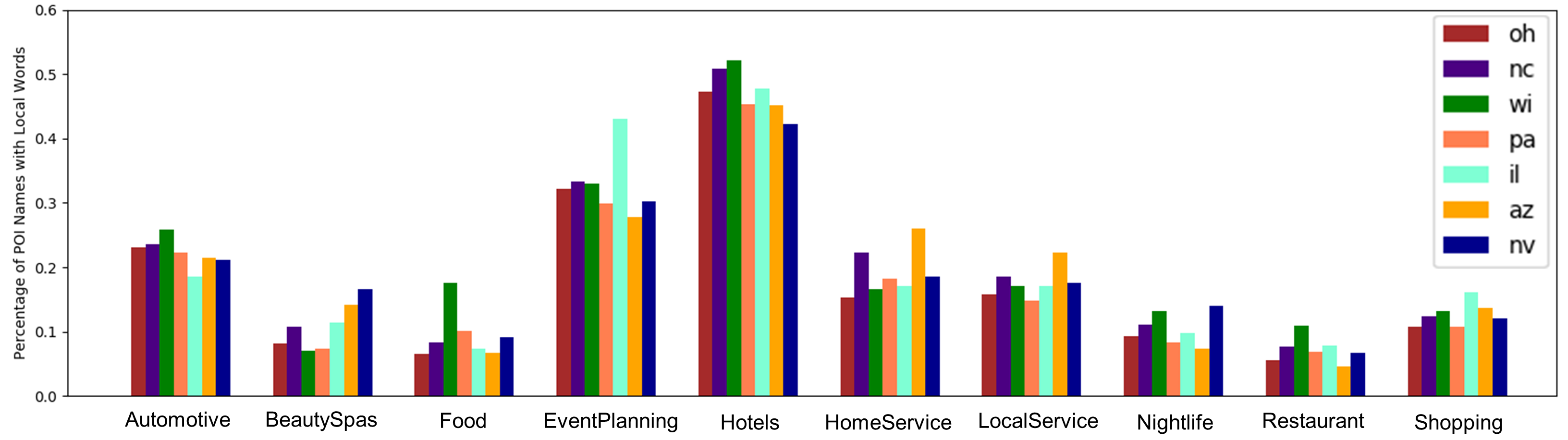}
	\caption{The percentages of POI names that contain local terms across POI types and different metropolitan areas.} \label{word_local_usage}
\end{figure}

In order to examine the interaction between POI names and POI types, we need to first divide the dataset based on POI types. Yelp has grouped their POIs into $23$ root categories which include \textit{Restaurants}, \textit{Shopping}, \textit{Food}, \textit{Hotels \& Travel}, and other categories. We make use of these Yelp POI categories, and the POIs in each metropolitan area are divided into subsets based on their categories. Yelp allows one POI to belong to multiple categories (e.g., one POI can be both \textit{Restaurants} and \textit{Nightlife}), and therefore the same POI is put into more than one subset when multiple categories exist. Not all metropolitan areas contain POIs in all $23$ categories. In addition, one metropolitan area may contain only a small number of POIs in a certain category, which can cause a biased result if those POIs are directly used for analysis. Thus, we only examine the POI types which are shared by all seven metropolitan areas and have at least one hundred POI instances in each area. Based on these criteria, we are left with ten categories, which are \textit{Automotive}, \textit{Beauty \& Spas},  \textit{Food}, \textit{Event Planning \& Services}, \textit{Hotels \& Travel}, \textit{Home Services}, \textit{Local Services}, \textit{Nightlife}, \textit{Restaurants}, and \textit{Shopping}. The TF-IDF weights from the first analysis are then re-used, and we extract the top 100 terms that have the highest TF-IDF weights in each metropolitan area and use them as the local terms. The percentage of POI names in each POI type that contain local terms is calculated using Equation \ref{local_usage}:
\begin{equation}
 \mathit{Pr}_{ij} = |\mathit{LP}_{ij}|/|P_{ij}| \label{local_usage}
\end{equation}
where $|\mathit{LP}_{ij}|$ is the number of POI names that contain any of the local terms in metropolitan area $i$ in POI type $j$, $|P_{ij}|$ is the total number of POI names in metropolitan area $i$ in POI type $j$, and $\mathit{Pr}_{ij}$ is the calculated percentage. The result is shown in Figure \ref{word_local_usage}. 

Two things can be observed in Figure \ref{word_local_usage}. First, there is an uneven usage of local terms across POI types. Overall, it seems that people (business owners) are more likely to include local terms in the names of hotels, event planning services, and automotive shops. In contrast, local terms are less likely to be used in the names of restaurants, shopping places, and beauty spas. This is understandable since we frequently see hotels (especially hotel chains) include city names as part of their names to indicate locations, such as \textit{holiday inn charlotte center city}. Meanwhile, restaurant names may focus on describing  food and cuisine styles to attract customers. Second, the uneven usage of local terms is highly consistent across the seven metropolitan areas. This result suggests that the identified local term usage patterns are not specific to a particular region but can be generalized to other geographic areas.


To quantify the similarity and difference of local term usage in different POI types across geographic regions, we employ Jensen-Shannon divergence (JSD), which measures the similarity between two probability distributions. Equation \ref{jsd_1} and \ref{jsd_2} show the calculation of Jensen-Shannon divergence, where $\mathit{KLD}(P || Q)$ is the Kullback–Leibler divergence. The output of JSD is in $[0,1]$, with $0$ indicating that the two distributions are highly similar and $1$ suggesting that the two distributions are largely different.
\begin{eqnarray}
 \mathit{JSD}(P || Q) = \frac{1}{2} \mathit{KLD}(P || M) + \frac{1}{2} \mathit{KLD}(Q || M) \label{jsd_1} \\
\mathit{KLD}(P || Q) = \sum_{i} P(i) \ln \frac{P(i)}{Q(i)}  \label{jsd_2}
\end{eqnarray}

JSD requires the input probabilities to sum to $1$. To satisfy this criterion, we normalize the initial percentage values using Equation \ref{normal_equation}:
\begin{equation}
	\textit{NPr}_i = \frac{\textit{Pr}_i}{\sum_{j} \textit{Pr}_j}
	 \label{normal_equation}
\end{equation}
We then iterate through the seven metropolitan areas and calculate the pair-wise JSD, and finally calculate the average JSD value (there are in total 21 values). The obtained average JSD is $0.007$, suggesting that the local term usage in different POI types are highly similar across geographic regions. The findings in this subsection can help us select suitable POI types in future for building computational models. For example, in the task of place name disambiguation, we may choose to focus on the POI names of certain types, such as \textit{Hotels} and \textit{Automotive} rather than \textit{Restaurant} and \textit{BeautySpas}, to extract more local terms which can then be associated with the related place names.

\subsection{Analyzing POI name change with geographic distance}
In this third analysis, we examine the change of POI names with geographic distance. Many phenomena follow Tobler's First Law and show a distance decay effect. Do POI names, which reflect many underlying social and cultural processes, also show such an effect? Here, we look into the  \textit{collective similarity} of POI names between metropolitan areas, namely how the POI names in one area are overall similar or dissimilar to the POI names in another area. For instance, we may expect the Phoenix metropolitan area to have more similar POI names compared with the Las Vegas metropolitan area than with the Cleveland metropolitan area.

One major challenge for this analysis is how to measure the \textit{collective similarity} of POI names between metropolitan areas. We propose two approaches to achieve this goal. The first and a straightforward approach is to group POI names in the same metropolitan area into a bag of words. This is similar to the TF-IDF approach discussed in our first analysis. However, we use only term frequency here, since TF-IDF artificially exaggerates the importance of local terms. While such an exaggeration is desired for local term extraction, it distorts the true frequencies of terms in POI names and therefore is not used in this analysis. We also do not remove the repeating POIs as we did in the first analysis. In short, we try to keep the POI names and term frequencies as they are in the real world in order to objectively model their change with geographic distance. The terms used in the POI names in each metropolitan area are combined together into a vector. We will refer to this approach as \textit{count-based vector}. To formally define this approach, let $r_1$ and $r_2$ represent two geographic regions, and each region contains a set of POIs. We derive the vector for a geographic region by counting the frequencies of terms in POI names. A common vocabulary $V$ is constructed based on all the terms of the POI names in a dataset. Thus, the names of POIs in the two regions, $r_1$ and $r_2$, can be collectively represented as two vectors:
\begin{eqnarray}
 <{\ w}_{11},\ {\ w}_{12},\ \dots ,\ {\ w}_{1\left|V\right|}> \\
<{\ w}_{21},\ {\ w}_{22},\ \dots ,\ {\ w}_{2\left|V\right|}>
\end{eqnarray}
where $\left|V\right|$ represents the size of the vocabulary, and ${\ w}_{ij}$ represents the count of term $j$ used in the POI names in geographic region $i$. 


While the count-based vector approach is straightforward, it does not capture the semantic similarity between terms. For example, the terms \textit{kiku} and \textit{sakana} are both used for the names of sushi restaurants in the dataset. The count-based vector will treat the two terms as completely different with a similarity of zero. However, the fact that these two terms both co-occur with \textit{sushi} suggests there exists certain degree of similarity between them. \textit{Word2vec} \citep{mikolov2013distributed} is a model that has been found to effectively capture the semantic similarity between terms. It is a neural network model which learns \textit{embeddings} (low dimension vectors) for terms. In this work, we use the word2vec model to learn embeddings for metropolitan areas based on POI names. The embeddings are learned by predicting the terms used in POI names based on a given region (e.g., what terms are likely to be used for POI names if the region is \textit{Phoenix, AZ}). The embeddings are condensed vectors, and the POI names in $r_1$ and $r_2$ can be represented as the two vectors below:
\begin{eqnarray}
 <{\ u}_{11},\ {\ u}_{12},\ \dots ,\ {\ u}_{1\left|d\right|}> \\
<{\ u}_{21},\ {\ u}_{22},\ \dots ,\ {\ u}_{2\left|d\right|}>
\end{eqnarray}
where $d$ is the dimensionality of the embeddings, which can be decided empirically. In this analysis, we set $d= 300$ following the recommendation from the literature \citep{mikolov2013distributed}. $u_{ij}$ is a weight value learned from the POI dataset. The word2vec model aims to minimize the objective function in Equation \ref{negativesampling}:
\begin{equation}
 J\boldsymbol{=\ }-\ {\mathrm{log} \sigma \left({\boldsymbol{w}}^T_o\boldsymbol{r}\right)\ }-\ \sum^K_{k=1}{{\mathrm{log} \sigma \left(-{\boldsymbol{w}}^T_k\boldsymbol{r}\right)\ }} \label{negativesampling}
\end{equation} 
where $\boldsymbol{r}$ is the embedding of one geographic region, ${\boldsymbol{w}}_o$ is the embedding of a term that is used for the POI names in region $\boldsymbol{r}$, while ${\boldsymbol{w}}_k$ is the embedding of a term not used in region $\boldsymbol{r}$ (which serves as negative samples). $\sigma $ is a sigmoid function: $\sigma \left(x\right)=\ \frac{1}{1+\ e^{-x}}$.


With different geographic regions represented as vectors in the same dimension, cosine similarity can be employed to measure the similarity of two vectors (Equation \ref{cosine}). $s(r_1, r_2)$ is then used as the collective similarity between regions $r_1$ and $r_2$.
\begin{equation}
 s(r_1, r_2) =\ \frac{\sum^{d}_{j=1}{w_{1j}w_{2j}}}{\sqrt{\sum^{d}_{j=1}{w^2_{1j}}}\ \sqrt{\sum^{d}_{j=1}{w^2_{2j}}}} \label{cosine}
\end{equation}

We apply both the count-based approach and word2vec to the Yelp POI dataset to derive vectors for the seven metropolitan areas. The center point of each metropolitan area is derived by averaging the location coordinates of the POIs in that area. We then employ Vincenty's formulae \citep{vincenty1975direct}, which is based on the assumption of an oblate spheroid, to calculate the distance between two metropolitan areas. We then perform both Pearson's and Spearman's correlation to examine the relation between the collective similarity of POI names and the geographic distance of the corresponding metropolitan areas. Table \ref{distance_correlation} shows the correlation results. 
\begin{table}[h]
	\centering
	\caption{Pearson and Spearman correlation coefficients between the collective similarity of POI names and geographic distance.}
	\label{distance_correlation}
	\begin{tabular}{|l|c|c|}
		\hline
		& \begin{tabular}[c]{@{}c@{}}Count-based  vector\end{tabular} & word2vec  \\ \hline
		Pearson  & -0.612 (p \textless  0.01)                                                      & -0.963 (p \textless  0.001) \\ \hline
		Spearman & -0.626 (p \textless  0.01)                                                      & -0.917 (p \textless  0.001) \\ \hline
	\end{tabular}
\end{table}
Overall, the collective similarity of POI names negatively and significantly correlates with geographic distance based on the four correlation coefficients in Table \ref{distance_correlation}, which suggests that POI names indeed \textit{gradually} become less similar with the increase of geographic distance. We emphasize \textit{gradually} here because either no change or abrupt change can lead to no correlation between POI name similarity and geographic distance. It is often natural to assume that place names at different locations are of course different, but our experiment result suggests that place names are not randomly different but follows a distance decay pattern. The statistical significance of the result is especially exciting given the fact that we have only 21 data points (21 region pairs from the seven metropolitan areas) for this correlation analysis. Such a result suggests that there is indeed a clear negative relation between POI name similarity and distance. In addition, it seems that word2vec better captures the POI name changes with geographic distance, as demonstrated by the higher correlation coefficients and stronger significances. 

\begin{figure}[h]
	\centering
	\includegraphics[width=\textwidth]{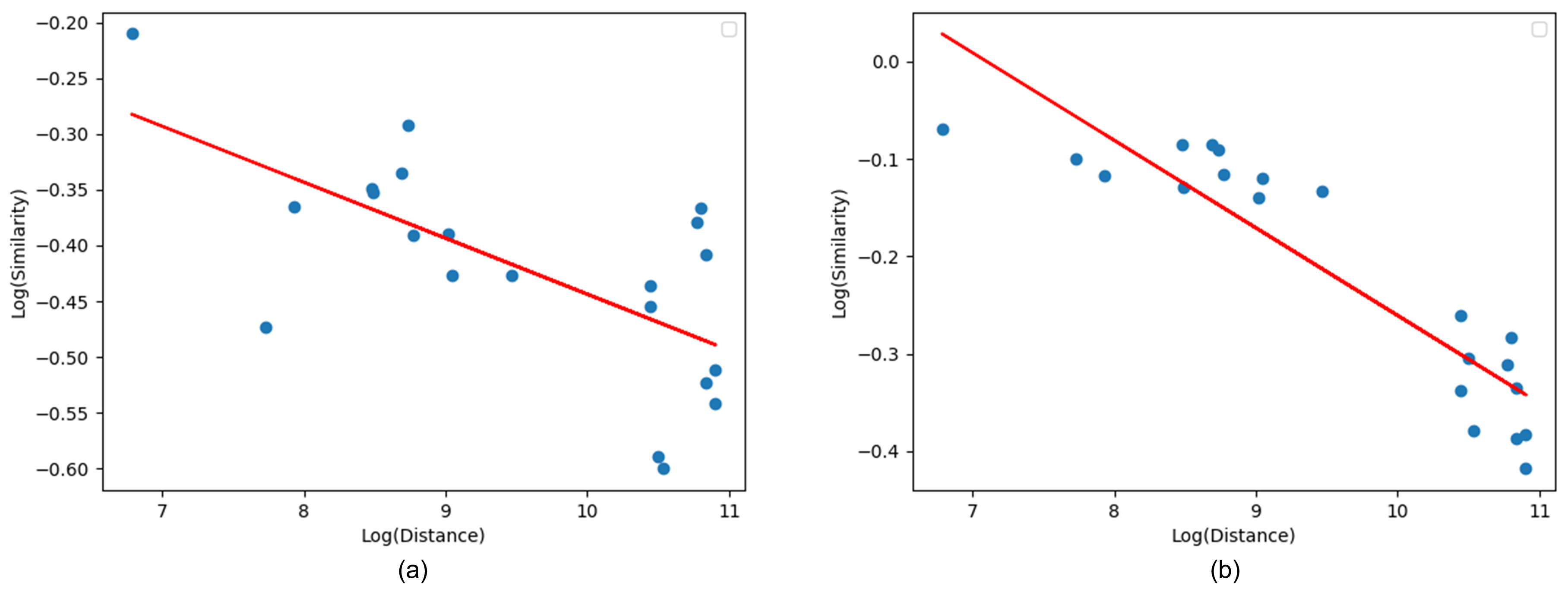}
	\caption{Fitting the collective similarity of POI names with geographic distance: (a) count-based vector; (b) word2vec.} \label{distance_decay}
\end{figure}

To further quantify the distance decay effect, we use a model $s = A * \frac{1}{d^\beta}$ to fit our data. We first transform it into its logarithmic form:
\begin{equation}
	\log s = A + \beta * \log d
\end{equation}
where $s$ is the collective similarity of POI names between two metropolitan areas, $A$ is a constant, $\beta$ is the slope, and $d$ is the geographic distance between them. We fit a linear regression model based on the logged values. Figure \ref{distance_decay} shows the result. 
In the count-based vector approach, we obtained an R-squared value $0.434$ and a slope of $-0.050$. Using word2vec, we obtained a R-squared value $0.828$ and a slope of $-0.090$. More credibility can be given to the result from word2vec since it better captures the semantic similarity between terms in POI names. A slope of -0.090 indicates there is a clear distance decay effect with the increase of geographic distance. Besides, it is interesting to see how the data points clearly fall in two groups in Figure \ref{distance_decay}(b), which is consistent with their geographic distributions shown in Figure \ref{POI_city} (a group of city pairs has closer geographic distances, while the other group of city pairs has farther geographic distances). It would be interesting to examine the POI names in more metropolitan areas to see if their POI names also follow the general trend along the red line in Figure \ref{distance_decay}(b).

\begin{figure}[h]
	\centering
	\includegraphics[width=\textwidth]{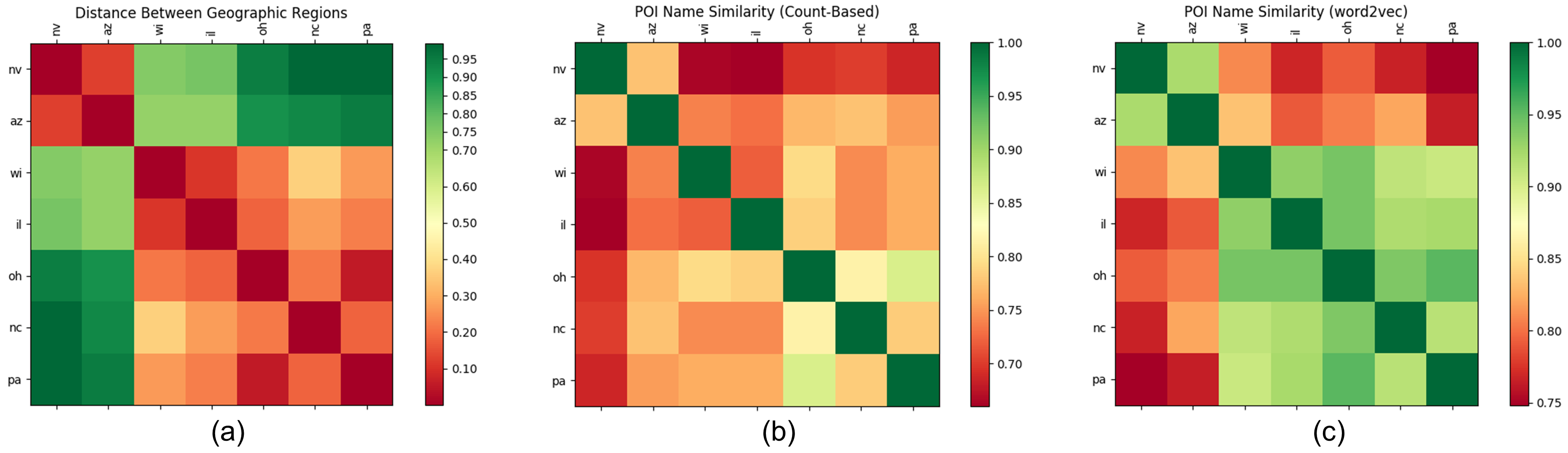}
	\caption{(a) The geographic distances between the seven metropolitan areas; (b) collective similarities  based on count-based vector; (c) collective similarities based on word2vec.} \label{similarity_matrix}
\end{figure}

To further examine the result difference between the count-based vector and word2vec, Figure \ref{similarity_matrix} shows the matrices of the geographic distances and the collective similarities obtained using the two approaches. It can be seen that the similarity pattern obtained using word2vec in sub figure (c) is closer to the distance pattern in sub figure (a) compared with the pattern from the count-based vector in sub figure (b). This result is consistent with the distance decay pattern observed in Figure \ref{distance_decay}. 

%% file: conclusion.tex
\section{Conclusions and future work}

Place names are texts given by people to natural or man-made geographic features. The act of assigning a name to space signifies the important moment of space and human experience integration, and further enhances the social construct of \textit{place}. Place names, as \textit{city-text}, reveal a considerable amount of information about the culture, lifestyle, community, and many other aspects of a city. While place names have long intrigued geographers, existing research often focuses on case-by-case qualitative descriptions related to the etymology or taxonomy of place names, or only considers place names as identifiers without analyzing their term usage and their relations with geographic distances.

This paper presents an empirical study on place names and their change with geographic distance. This study is based on an open dataset from Yelp, and examines more than $110,000$ POIs, such as restaurants, hotels, and local services, in seven metropolitan areas in the United States. We perform an exploratory analysis on the frequencies of terms used in POI names, and find the term usage follows Zipf's law. We further conduct three analyses focusing on \textit{place names}, \textit{place types}, and \textit{spatial footprints} respectively. We adapt the technique of term frequency and inverse document frequency in geographic context to identify local terms, and examine the term usage in the POI names in different types of POIs. We find an uneven usage of local terms across POI types (e.g., auto repairs are more likely to use local terms than restaurants), and such a usage pattern is highly consistent across different geographic regions. Finally, we test two approaches, count-based vector and word2vec, to model the collective similarity of POI names in different regions, and find a distance decay effect in the collective similarity of POI names. 

This work is only a first step towards quantitatively and systematically examining place names and their relations with geographic distances. A number of topics can be explored in the near future. First, all the analyses are conducted based on the seven metropolitan areas available in the Yelp dataset. While a large number of POI names are examined, it would be interesting to apply the analyses to more metropolitan areas in other regions (e.g., north west and mid-south) as well as within local regions  to further test the findings from this work.
Second, we have so far used whole terms for the analyses, and it would be interesting to examine the parts or chunks of a term for measuring the collective similarity of place names. For example, the place names, \textit{Wauwatosa} in Wisconsin, \textit{Wawatasso} in Minnesota, and \textit{Wahwahtaysee} in Michigan, share similar chunks, and may have higher similarity values when a chunk-based approach is used.
Third, future research can be conducted on how to integrate the information extracted from place names with existing computational models for tasks such as place name disambiguation. While Wikipedia articles and other datasets have been frequently used for training place-based models, there are situations when we have only short Wikipedia descriptions or no description for places. Local information extracted from place names can serve as additional resources to improve existing models.